\documentclass[letterpaper, 10 pt, conference]{ieeeconf}  

\IEEEoverridecommandlockouts                              

\usepackage{amsmath} 
\usepackage{amssymb}  
\usepackage{gensymb}
\usepackage{bm}

\usepackage{graphicx}
\usepackage{color}
\usepackage{import}
\usepackage[font=footnotesize]{caption} 
\usepackage{subcaption}
\usepackage{hyperref}
\usepackage{cite}
\usepackage{algorithm}
\usepackage{algpseudocode} 

\title{\LARGE \bf
A Front-End for Dense Monocular SLAM using a Learned Outlier Mask Prior}

\author{Yihao Zhang and John J. Leonard
\thanks{Yihao Zhang and John Leonard are with the Computer Science and Artificial Intelligence Laboratory, Massachusetts Institute of Technology, Cambridge, MA 02139 USA {\tt\small \{yihaozh, jleonard\}@mit.edu}}%
\thanks{This work was supported by ONR MURI grant N00014-19-1-2571,
ONR grant N00014-18-1-2832, and ARPA-E award
DE-AR0001218 under the DIFFERENTIATE Program.}} 

\begin{document}

\maketitle
\thispagestyle{empty}
\pagestyle{empty}

\begin{abstract}

Recent achievements in depth prediction from a single RGB image have powered the new research area of combining convolutional neural networks (CNNs) with classical simultaneous localization and mapping (SLAM) algorithms. The depth prediction from a CNN provides a reasonable initial point in the optimization process in the traditional SLAM algorithms, while the SLAM algorithms further improve the CNN prediction online. However, most of the current CNN-SLAM approaches have only taken advantage of the depth prediction but not yet other products from a CNN. In this work, we explore the use of the outlier mask, a by-product from unsupervised learning of depth from video, as a prior in a classical probability model for depth estimate fusion to step up the outlier-resistant tracking performance of a SLAM front-end. On the other hand, some of the previous CNN-SLAM work builds on feature-based sparse SLAM methods, wasting the per-pixel dense prediction from a CNN. In contrast to these sparse methods, we devise a dense CNN-assisted SLAM front-end that is implementable with TensorFlow and evaluate it on both indoor and outdoor datasets.

\end{abstract}

\section{Introduction}
\label{sect:intro}

Simultaneous localization and mapping (SLAM) has been a long-standing problem~\cite{cadena2016past}. In particular, monocular SLAM requires simultaneously solving for the depths of points in the images and the poses of the camera along its trajectory given only the sequence of images that the camera has captured as measurements. The problem is difficult because the data association between a pixel in a frame and the corresponding pixel in a second frame is also unknown along with the depths and the camera poses. Advances in computer vision and the recent success in the area of single-image depth prediction with CNNs \cite{laina2016deeper, godard2017unsupervised, zhou2017unsupervised, mahjourian2018unsupervised, bian2019unsupervised, godard2019digging} have brought us exciting new opportunities in monocular SLAM. The depth prediction from a CNN can provide a reasonable guess for one of the unknowns in the SLAM problem, the depth, helping to eliminate this unknown from the problem. Recent monocular SLAM methods that take advantage of the learned depth map often use it as an initial point in the depth and pose optimization scheme \cite{tateno2017cnn, loo2018cnn, yang2018deep, yang2020d3vo, bloesch2018codeslam, zhi2019scenecode, czarnowski2020deepfactors}. They have shown surpassing performance over the conventional SLAM methods in their accuracy and their robustness to ill conditions (e.g. pure rotations and textureless environments) \cite{tateno2017cnn, loo2018cnn, yang2018deep}.

\setlength{\belowcaptionskip}{-20pt}
\begin{figure}[t]
\vspace{6pt} 
    \begin{center}
    \centering
    \begin{subfigure}{0.48\linewidth}
    \centering
		\includegraphics[width=1.0\columnwidth]{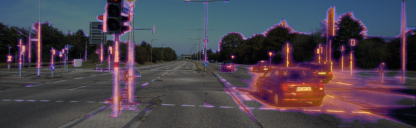}%
	\label{fig:orig_src}
	\end{subfigure}%
	\hspace{1pt}
	\begin{subfigure}{0.48\linewidth}
	\centering
		\includegraphics[width=1.0\columnwidth]{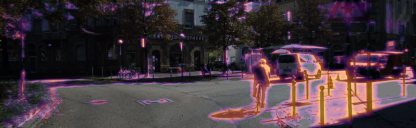}%
	\label{fig:orig_l1}
	\end{subfigure}%
	\par\smallskip
	\centering
    \begin{subfigure}{0.48\textwidth}
        \centering
        \includegraphics[width=0.99\textwidth]{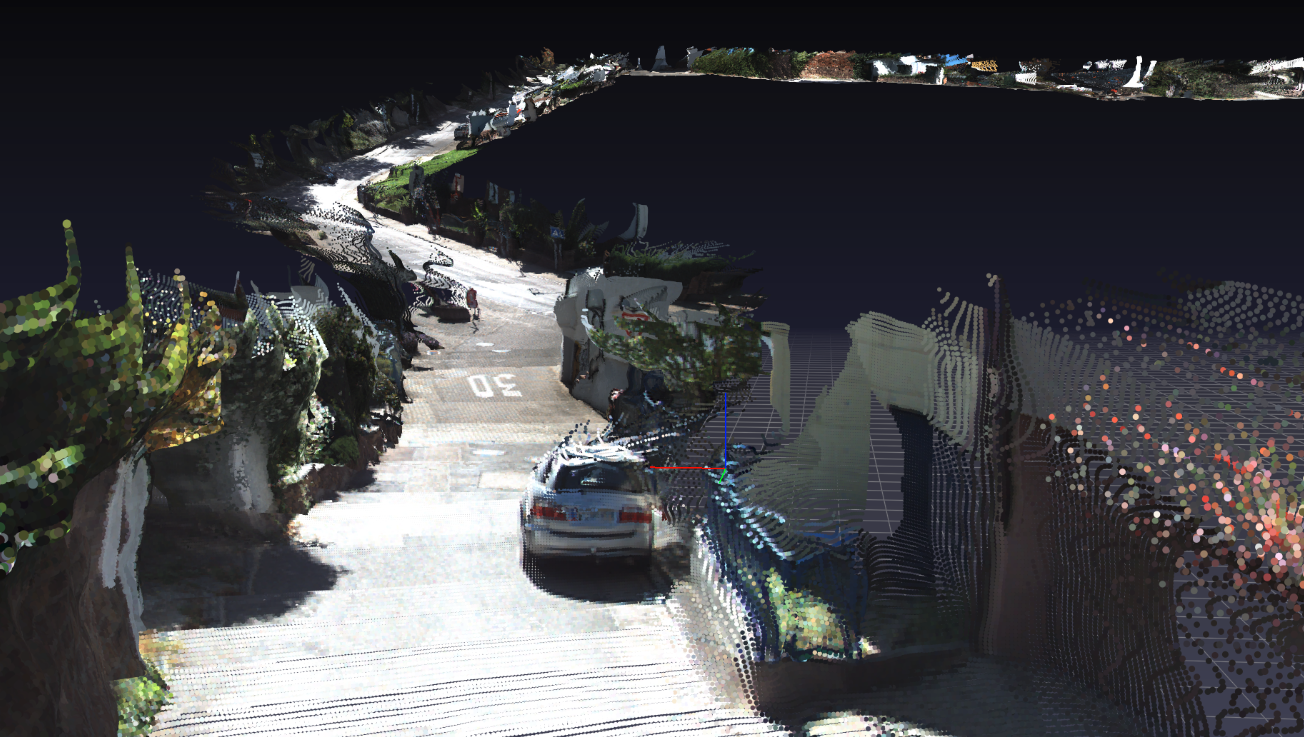}
	\end{subfigure}%
	\end{center}
    \caption{Top row: The learned outlier mask. The brighter and yellower regions indicate the predicted outliers in the photometric consistency due to dynamic objects and static occlusions. Bottom row: A dense point cloud of KITTI \cite{Geiger2013IJRR} odometry sequence 10 generated by our SLAM front-end. (Far points are excluded for cleaner visualization.)}
\label{fig: pt_cloud_intro}
\end{figure}
\setlength{\belowcaptionskip}{0pt} 

A number of these CNN-SLAM approaches \cite{tateno2017cnn, loo2018cnn, yang2018deep, yang2020d3vo} build upon existing mature SLAM systems and integrate the CNN depth prediction into those systems. These methods show high reliability due to their foundation on tested SLAM technologies. In the meantime, they also inherit the underlying map representation, which is typically a sparse representation \cite{loo2018cnn, yang2018deep, yang2020d3vo}, from the SLAM systems that they build upon.

Some recent work has looked beyond merely utilizing CNNs to initialize the depth estimates. The line of work \cite{bloesch2018codeslam, zhi2019scenecode, czarnowski2020deepfactors} following CodeSLAM \cite{bloesch2018codeslam} experimented reducing the depth map onto a lower dimensional manifold by using the learned CNN decoder for the purpose of efficient joint optimization on multiple frames. D3VO \cite{yang2020d3vo} also started to incorporate not only the learned depth but also the learned depth uncertainty into the SLAM pipeline.

We present a dense monocular CNN-assisted SLAM front-end that is fully implementable with TensorFlow. The focus is on the pose estimation performance without the back-end inference. We show that when the CNN training objective is aligned with the online pose estimation objective, the outlier mask, a by-product learned from the training process can be used together with a classical probability model to improve the pose estimation accuracy through better handling the outliers due to static occlusions and dynamic objects. As semantic reconstruction becomes more and more important for visual systems, our SLAM front-end also processes the semantic segmentation images using forward propagation to improve their quality. We evaluate our method on both indoor and outdoor datasets.

\section{Related Work}
\label{sect:related-work}

\subsection{Classical Monocular SLAM Methods}

There exist two branches of monocular SLAM methods, feature-based methods and direct methods. Represented by ORB-SLAM \cite{mur2017orb}, the pose estimation in feature-based methods is done by first finding the corresponding points between two frames and then computing their relative pose given the correspondences. These methods create a sparse map since only a few hundred key points in an image are involved in the computation, while all the other pixels are discarded. Direct methods \cite{newcombe2011dtam, engel2013semi, engel2014lsd, engel2017direct} rely on direct image alignment, which is based on photometric consistency, for pose estimation. These methods can be used to construct a sparse, semi-dense, or dense map depending on the choice of points of interest. The tracking module in a typical direct visual odometry pipeline \cite{engel2014lsd} tracks the current frame against a previously established keyframe which is associated with a depth map initialized by given statistics (e.g. average scene depth). The tracking step commonly includes performing direct image alignment for pose estimation between the current frame and the keyframe and updating the keyframe depth map given the estimated pose. In this work, we leverage several techniques from the previous work in direct methods with modifications to build our dense SLAM front-end pipeline.

\subsection{SLAM using Learned Depth}

The first work of combining SLAM with CNN depth prediction is CNN-SLAM \cite{tateno2017cnn}, where the predicted depth images from a CNN initialize the keyframe depth estimates in LSD-SLAM (large-scale direct monocular SLAM) \cite{engel2014lsd}, replacing the random initialization in the original LSD-SLAM. Their results showed that monocular SLAM armed with the CNN depth prediction could better recover the absolute scale of a scene, densify the depth estimates in texture-less regions, and deal with degenerative camera motions. CNN-SVO \cite{loo2018cnn} shared the same idea with CNN-SLAM but instead of LSD-SLAM it was based on SVO (semi-direct monocular visual odometry) \cite{forster2014svo}.

Later CNN-SLAM approaches become more sophisticated in the use of CNNs. DVSO \cite{yang2018deep}, which is built upon monocular DSO (direct sparse odometry) \cite{engel2017direct}, forms a virtual stereo loss term in the sparse bundle adjustment using the predicted left and right stereo disparity images so that the CNN depth prediction does not only provide an initial point at the start of the optimization but also strengthens the quality of the optimized depth map through the stereo loss term. CodeSLAM \cite{bloesch2018codeslam} uses a variational auto-encoder (VAE) \cite{kingma2013auto} to learn a lower-dimensional latent representation of depth images. It updates the lower-dimensional latent representation instead of the depth image itself to achieve efficient optimization of the dense depth image. SceneCode \cite{zhi2019scenecode} and DeepFactors \cite{czarnowski2020deepfactors} extends CodeSLAM by incorporating an additional semantic consistency loss and a feature-based re-projection loss to further improve the system performance.

DVSO \cite{yang2018deep} and CodeSLAM \cite{bloesch2018codeslam} explored the utility of CNNs in different ways, not only using them for depth estimate initialization. We present a third way that is fusing a learned outlier mask as a prior into a classical probability model to handle outliers in the photometric consistency loss caused by dynamic objects and static occlusions in the environment. Our method can be an add-on to their approaches.

\subsection{Depth Prediction from a Single Image}

CNNs can be trained in either a supervised manner \cite{laina2016deeper} or an unsupervised manner \cite{godard2017unsupervised, zhou2017unsupervised, mahjourian2018unsupervised, bian2019unsupervised, godard2019digging} to enable depth prediction. We are interested in unsupervised training with monocular video \cite{zhou2017unsupervised, mahjourian2018unsupervised, bian2019unsupervised, godard2019digging}. We found that the outlier mask, a by-product from this training process, could be engineered into a SLAM system to improve its tracking performance. 

\section{Preliminaries}
\label{sect:prel}

\subsection{Photometric Consistency Loss}

At the core of direct methods (e.g. \cite{engel2017direct}) and unsupervised learning of depth approaches (e.g. \cite{zhou2017unsupervised}) is the photometric consistency which states that if a pixel in one frame and a pixel in another frame correspond to the same 3D point, they should have the same intensity in the images. Therefore, the photometric consistency loss is defined in our work as:
\begin{gather}
    L_{pho}(I', I) = \frac{1}{|V|}\sum_{p\in{\Omega(I)}}{W(p)\|I'(\Pi(p)) - I(p)\|_m}
    \label{eq:photoloss} \\
    \Pi(p) = \Pi(p, D(p), R, t) = \left[ {\begin{array}{c}u' \\ v' \end{array} } \right]
    \label{eq:proj} \\
    d'\left[ {\begin{array}{c}u' \\ v' \\ 1\end{array} } \right] = K\left[ {\begin{array}{cc}R & t\end{array} } \right]homo\biggl(K^{-1}D(p)\left[ {\begin{array}{c}u \\ v \\ 1\end{array} } \right]\biggr)
    \label{eq:depth}
\end{gather}
where $I$ and $I'$ are two nearby frames, $p = \left[ {\begin{array}{cc}u&v\end{array} } \right]^{T}\in{\Omega(I)}$ is a pixel in the image space $\Omega(I)$, $W(p)$ is the weight applied to the loss term at pixel $p$, $V$ is the set of valid pixels projected within the image boundaries, $\|\cdot\|_m$ is the m-norm, and $\Pi(p)$ is the perspective projection function, which is expressed in detail in (\ref{eq:depth}) where $D(p)$ is the depth at pixel $p$, $R$ and $t$ are the rotation matrix and the translation vector from $I$ to $I'$, $K$ is the 3-by-3 camera instrinsic matrix, $homo$ denotes transforming to the homogeneous coordinates, and $d'$ is the projected depth in frame $I'$. The 1-norm is typically used in CNN training for robustness to outliers, while the 2-norm is standard in direct visual odometry methods. $\Pi(p)$ yields non-integer values, which are inappropriate to index pixels, and therefore we use bilinear interpolation to interpolate between pixels. The bilinear interpolation also retains the differentiability of the photometric loss for use in gradient-based optimization methods. In addition, the projected pixel $\Pi(p)$ can fall outside the image boundaries. We exclude these invalid pixels and normalize the photometric loss by the number of valid pixels $|V|$, as done in \cite{bian2019unsupervised}. The photometric loss in (\ref{eq:photoloss}) can be easily extended to RGB-channels often used in CNN training.

\subsection{Learning Depth and Outlier Mask from Video}
\label{sect: learn_mask}

Prior work \cite{zhou2017unsupervised, mahjourian2018unsupervised, bian2019unsupervised} has discussed in detail how to learn depth and ego-motion from monocular video. We will focus on discussing the outlier mask. The closest definition of the outlier mask in our work is the explainability mask in \cite{zhou2017unsupervised}. For each pixel in the image, an outlier mask value within $[0, 1]$ is predicted. This value serves as the weight term $W(p)$ in (\ref{eq:photoloss}) during training. Intuitively, if a pixel corresponds to a 3D point on a dynamic object or on an occlusion surface, being an outlier, the photometric consistency will be violated at that pixel, so its photometric consistency loss should not be counted and $W(p)$ will tend to zero. In order to avoid the trivial case of the learned outlier mask being assigned to zero for every pixel during training, a cross-entropy loss with an all-ones mask is employed as regularization \cite{zhou2017unsupervised}. Unlike \cite{zhou2017unsupervised} where the outlier mask is predicted from the ego-motion network given the two frames for pose prediction, our model ties the mask prediction to the depth network. In this way, the mask can be predicted from a single image. The detailed training procedures that we followed are in \cite{zhang2021bootstrapped}.


\section{Method}

The main system pipeline is described in this section. An overview of the pipeline is given in Algorithm \ref{alg: overview}. The system is a SLAM front-end. It follows the paradigm of alternating between mapping and tracking \cite{klein2007parallel}. The final outputs are an odometry chain, the depth maps and semantically segmented images for the keyframes. The detailed computing procedures are illustrated in the following subsections.

\begin{algorithm}[H] 
\caption{System Overview}
\label{alg: overview}
\begin{algorithmic} 
\renewcommand{\algorithmicrequire}{Input:} 
\renewcommand{\algorithmicensure}{Output:}
\Require image sequence $\{I_i\}$, $i = 1, ..., n$
\Ensure camera poses $\{T_i\}$ for all frames, depth maps $\{D_j\}_{kf}$ and semantically segmentated images $\{S_j\}_{kf}$ for keyframes, $\{j\} \subseteq \{1, ..., n\}$
\Statex
\Function{insert a keyframe }{$I$}
    \State {Predict the depth map, outlier mask, and semantic \\ \hspace{0.43cm} segmentation for $I$.}
    \State {Initialize the probability model parameters $\alpha$, $\beta$, $\mu$, \\ \hspace{0.415cm} and $\sigma$ for each pixel in $I$ using the predicted depth \\ \hspace{0.415cm} map and outlier mask.}
\EndFunction
\Statex
\State \Call {insert a keyframe }{$I_1$}
\For {i = 2 to n}
    \State {Estimate $T_i$ for $I_i$ against the keyframe.}
    \State {Update the depth map of the keyframe given $T_i$.}
    \State {Update $\alpha$, $\beta$, $\mu$, and $\sigma$ for each pixel in the keyframe.}
    \If {the keyframe criteria are satisfied}
        \State \Call {insert a keyframe }{$I_i$}
        \State {Fuse the semantic class probabilities between the \\ \hspace{0.93cm} last keyframe and this keyframe.}
    \EndIf
\EndFor
\end{algorithmic}
\end{algorithm}

\subsection{Keyframe Insertion}
\label{sect: kf_insert}

A frame is inserted as a keyframe when the system initializes on the first frame of the sequence or when the frame has passed the keyframe criteria (\ref{sect: kf_check}). A keyframe is associated with multiple attributes, a pose, a color image, a gray-scale image, a semantic segmentation image, a depth map, an outlier mask, and a map of the parameters in the probability model that we will use. Upon the keyframe insertion, the color image is fed into our CNN models to predict the semantic segmentation image, the depth map, and the outlier mask. The outlier mask and the depth map are used to compute the prior distributions for the probability model illustrated next.

Suppose we are at time step $k$ and have collected a depth measurement $d_k$ for a pixel. We can model the distribution on this depth measurement as the following \cite{pizzoli2014remode, vogiatzis2011video}:
\begin{equation}
    p(d_k|\hat{d},\rho) = \rho\mathcal{N}(d_k|\hat{d},\tau^{2}_{k}) + (1 - \rho)\mathcal{U}(d_k|d_{min}, d_{max})
    \label{eq: p_model}
\end{equation}
where $\rho \in [0, 1]$ is the inlier probability of that measurement, $\hat{d}$ is the true depth, $\tau^{2}_{k}$ is the variance for an inlier depth measurement in a Gaussian probability model, $[d_{min}, d_{max}]$ is the interval where an outlier depth measurement is uniformly distributed. Essentially, we assume if the depth measurement is an inlier, it follows a Gaussian distribution, and otherwise, it should follow a uniform distribution. The inlier ratio $\rho$ is modeled to follow a Beta distribution parameterized by $\alpha$ and $\beta$. This model (\ref{eq: p_model}) is used for depth estimate fusion in \cite{vogiatzis2011video} assuming camera poses are given a priori. Evidence of why this model is sound can be found in \cite{vogiatzis2011video}.

To perform the posterior update using this probability model (i.e. to compute $p(\hat{d},\rho|d_1, ..., d_k)$), we need a prior Gaussian distribution on $\hat{d}$ and a prior Beta distribution on $\rho$. For the Beta prior parameterized by $\alpha_0$ and $\beta_0$, we set the ratio of $\alpha_0$ to $\alpha_0+\beta_0$, which is the Beta mean $\mathbb{E}{[\rho]}$, to be the predicted outlier mask value. We further set $\alpha_0+\beta_0$ as a fixed tuning parameter in order to compute the values of $\alpha_0$ and $\beta_0$. This tuning parameter essentially controls the variance of the Beta distribution.

The prior Gaussian distribution has its mean $\mu_0$ set equal to the depth prediction from the CNN and its standard deviation $\sigma_0$ set equal to a percentage of the predicted depth. This percentage is set as another tuning parameter. Alternatively, the standard deviation can be predicted by a CNN using the concept of heteroscedastic aleatoric uncertainty \cite{kendall2017uncertainties, yang2020d3vo}. However, this would introduce extra complexity in the coupling between the SLAM system and the CNN, making the network training process difficult. We thus opted not to use a predicted standard deviation.

The prior parameters, $\alpha_0$, $\beta_0$, $\mu_0$, and $\sigma_0$, will be updated in the posterior computation (\ref{sect: post_update}). We denote the updated parameters at the current time step as $\alpha_k$, $\beta_k$, $\mu_k$, and $\sigma_k$.

\subsection{Pose Estimation}
\label{sect: pose_est}

After the keyframe has been established, subsequent frames are tracked against the keyframe. The pose of a frame with respect to the keyframe is estimated with direct image alignment \cite{engel2017direct, engel2014lsd}:
\begin{equation}
    \min_{T, a, b}{\left\{L_{pho}(I_{f}, aI_{kf} + b) + w[(a - 1)^{2} + b^{2}]\right\}}
    \label{eq: pose_est}
\end{equation}
where $T = [R, t]$ is the pose matrix, $a$ and $b$ are the affine lighting transformation coefficients \cite{engel2017direct}, and $I_{f}$ and $I_{kf}$ are the gray-scale images of the current frame and the keyframe. The second term in (\ref{eq: pose_est}) is the regularization on the affine lighting coefficients \cite{yang2020d3vo}. This regularization loss is weighted by $w$. The depth map involved in the computation of $L_{pho}$, as seen in (\ref{eq:proj}), consists of the Gaussian mean $\mu_{k-1}$ for every pixel. The weight $W(p)$ used in the computation, as seen in (\ref{eq:photoloss}), is the mean inlier probability $\mathbb{E}{[\rho]}$ which is the ratio of $\alpha_{k-1}$ to $\alpha_{k-1}+\beta_{k-1}$ for pixel $p$. In other words, we down-weight the photometric loss at a pixel if the pixel is likely to be an outlier according to our probability model.

The optimization in (\ref{eq: pose_est}) is solved by Newton's method to take advantage of the built-in Hessian matrix computation in TensorFlow. To further reduce the outlier effects, we use the Huber loss function, implemented as iteratively re-weighted least squares, for the photometric loss $L_{pho}$. During the iterative optimization, the increments are accumulated on the $SE(3)$ pose matrix \cite{engel2017direct}.

The pose estimation is iteratively performed on an image pyramid which accommodates the resolution of the image and the camera motion aggressiveness. In addition, the initial pose estimate at the beginning of the iteration is computed by a constant motion model given the past pose estimates \cite{engel2017direct}.

\subsection{Discrete Depth Search}
\label{sect: depth_search}

After the pose is estimated, we re-estimate the depth map $D_{kf}$ associated with the keyframe given the estimated pose and the estimated affine lighting transformation coefficients $a$ and $b$. For each pixel in $I_{kf}$, we maximize the normalized cross-correlation (NCC) between $I_{kf}$ and $I_f$ over a $3 \times 3$ patch ($\omega$) around that pixel: 
\begin{equation}
    \max_{D_{kf}(\omega)}{\frac{\sum_{p\in{\omega}}{I'_{kf}(p)I_f(\Pi(p))}}{\sqrt{\sum_{p\in{\omega}}{I'_{kf}(p)^2}}\sqrt{\sum_{p\in{\omega}}{I_f(\Pi(p))^2}}}}
    \label{eq: depth_update}
\end{equation}
where $I'_{kf} = aI_{kf} + b$. This optimization is solved by the simple gridding method. For each pixel in $\omega$, the depth range of two standard deviations ($2\sigma_{k-1}$) above and below the Gaussian mean $\mu_{k-1}$ is discretized into a few points. We discretize the entire patch ($D_{kf}(\omega)$) together to be compatible with the batch processing in TensorFlow. The center pixel is assigned with the discrete depth point that maximizes the NCC score (\ref{eq: depth_update}). This re-estimated depth is treated as a depth measurement ($d_k$) in the posterior update. 

\subsection{Posterior Update}
\label{sect: post_update}

The posterior given our measurement probability model in (\ref{eq: p_model}) is non-trivial, but the authors of \cite{vogiatzis2011video} showed that the posterior can be approximated as the product of a Beta distribution and a Gaussian distribution:
\begin{equation}
    q(\hat{d},\rho) = Beta(\rho|\alpha_{k},\beta_{k})\mathcal{N}(\hat{d}|\mu_{k},\sigma^{2}_{k})
\end{equation}
Therefore, computing the posterior only involves updating the four parameters, $\alpha_k$, $\beta_k$, $\mu_k$, and $\sigma_k$ from their values at time step $k-1$. The detailed update method can be found in the supplementary material of \cite{vogiatzis2011video}. During the update, the depth measurement standard deviation $\tau_k$ in (\ref{eq: p_model}) is needed. In \cite{pizzoli2014remode}, this measurement uncertainty is obtained by assuming a standard deviation of one pixel in the image during the depth search step and back-projecting this one pixel uncertainty to depth uncertainty using geometry. We instead use the simple approximation found in \cite{engel2013semi} to back-propagate the one pixel uncertainty:
\begin{equation}
    \tau^{2}_{k} = (\frac{\delta_{d}}{\delta_{\lambda}})^2\tau_{\lambda}^2
\end{equation}
where $\tau_{\lambda}$ is the assumed one-pixel standard deviation, $\delta_{d}$ is the depth search range in \ref{sect: depth_search}, and $\delta_{\lambda}$ is the corresponding pixel search range along the epipolar line in image $I_f$. In this way, we back-propagate the uncertainty using a numerically approximated Jacobian of the projection function.

\subsection{Keyframe Criteria and Keyframe Propagation}
\label{sect: kf_check}

The steps described in \ref{sect: pose_est} -- \ref{sect: post_update} are repeated for each incoming frame until the current frame is too far from the keyframe. We use two criteria to decide whether to insert the current frame as a new keyframe. First, we set a threshold for the maximum number of frames allowed to pass after a keyframe. Second, we compute the number of valid pixels projected within the image boundaries, which is $|V|$ in (\ref{eq:photoloss}), in the last iteration of the pose estimation step (\ref{sect: pose_est}). If the percentage of valid pixels is below a threshold, we insert a new keyframe. More sophisticated criteria can be found in \cite{engel2017direct}. However, the two criteria mentioned here are enough for us to evaluate our front-end.

After a frame is chosen to be a keyframe, we proceed back from the keyframe insertion step (\ref{sect: kf_insert}). The estimated pose of the frame is stored as the keyframe pose. The semantic labels in the form of class probabilities for pixels that have high inlier probabilities in the last keyframe are propagated onto the new keyframe using the perspective projection function (\ref{eq:proj}) with the latest depth map estimate and the relative pose estimate. The propagated class probabilities are fused with the newly predicted softmax probabilities \cite{mccormac2017semanticfusion}:
\begin{equation}
    P(c|I_{0,...,j}) = \frac{1}{Z}P(c|I_{0,...,j-1})P(O = c|I_j)
\end{equation}
where $j$ indexes keyframes, $P(O = c|I_j)$ is the predicted softmax probability at a pixel for class $c$, $P(c|I_{0,...,j-1})$ is the class probability propagated from the last keyframe, and $P(c|I_{0,...,j})$ is the fused class probability.

\section{Results}

We evaluate our SLAM front-end on the KITTI dataset \cite{Geiger2013IJRR} and the ScanNet dataset \cite{dai2017scannet}. KITTI is an outdoor driving dataset commonly used for evaluating odometry systems, whereas the indoor dataset, ScanNet, provides various ground truth information that is useful for evaluating intermediate system results. 

\subsection{Training on KITTI}

For the KITTI dataset, the prior work on unsupervised learning of depth provides high quality trained weights. As our network architecture is compatible with the one in \cite{mahjourian2018unsupervised}, we used their publicly released weights to initialize the weights in our depth network. The semantic segmentation network was trained on the Cityscapes dataset \cite{Cordts2016Cityscapes}. We followed \cite{zhang2021bootstrapped} to further train the depth network on Eigen et al.'s training split \cite{eigen2014depth} (a subset of KITTI raw data) to enable the outlier mask prediction. Our depth network and semantic segmentation network both have the DispNet architecture \cite{mayer2016large} for simplicity and run on $128 \times 416$ images.

\begin{figure}[t] 
\vspace{6pt} 
\begin{center}
	\centering
	\begin{subfigure}{0.49\linewidth}
		\includegraphics[width=1.0\columnwidth]{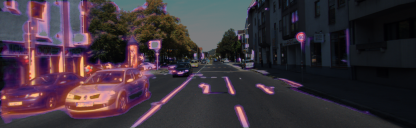}%
	\label{fig:orig_src}
	\end{subfigure}%
	\hfill
	\begin{subfigure}{0.49\linewidth}
		\includegraphics[width=1.0\columnwidth]{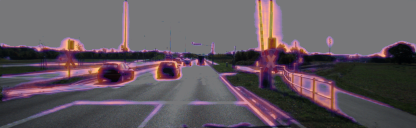}%
	\label{fig:ours1_src}
	\end{subfigure}%
	\par\smallskip
	\centering
	\begin{subfigure}{0.49\linewidth}
		\includegraphics[width=1.0\columnwidth]{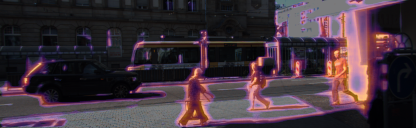}%
	\label{fig:ours1_l1}
	\end{subfigure}%
	\hfill
	\begin{subfigure}{0.49\linewidth}
		\includegraphics[width=1.0\columnwidth]{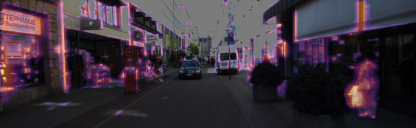}%
	\label{fig:ours2_l1}
	\end{subfigure}
\end{center}
\caption{Visualization of the predicted outlier mask. The darker regions indicate inlier pixels, whereas the brighter and yellower regions represent the predicted outlier pixels. Dynamic objects such as cars and people, and static occlusions such as poles and object boundaries are clearly identified.} 
\label{fig: outlier_mask}
\end{figure}

We provide some visualization of the outlier mask predicted from a single image in Fig. \ref{fig: pt_cloud_intro} and Fig. \ref{fig: outlier_mask}. As seen in the visualization, the mask is concentrated on dynamic objects, such as people and cars, and object boundaries, which are usually occlusions since the scene behind these boundaries is typically revealed in the next frame. There are also cases where the mask is on thin surface textures (e.g. the white lane markers on the road). These regions tend to have high photometric consistency error because it is difficult to project the pixels exactly to match these thin textures due to the errors in the predicted pose and depth. The network has learned to down-weight the photometric consistency loss at these difficult image regions using the mask. Overall, the outlier mask predicted from a single image looks similar to the explainability mask predicted from two nearby frames in \cite{zhou2017unsupervised}. It is reasonable to set the expected inlier probability $\mathbb{E}[\rho]$ in the prior to be the outlier mask value (\ref{sect: kf_insert}).

\setlength{\belowcaptionskip}{-15pt}

\subsection{Trajectory Evaluation on KITTI}

We evaluated the performance of our method on the KITTI odometry sequences 09 and 10 which were excluded in the network training set (for both our training and the training of the public checkpoint \cite{mahjourian2018unsupervised}). Our pipeline is only a front-end without any back-end smoothing, so drift is accumulated more quickly in our system than in a full SLAM system. To fairly evaluate our system, we chose the evaluation method in \cite{zhou2017unsupervised}, where the odometry system is run on 5-frame snippets, and the mean absolute trajectory error (ATE) is computed over all snippets. For each snippet, the first frame was inserted as the reference keyframe and the following four frames were tracked against that keyframe. We compare five different settings: (1) the full system, (2) the system without the posterior update, (3) with the posterior update but using an all-ones mask instead of the predicted outlier mask as the prior, (4) with the posterior update and the outlier mask prior but without down-weighting during the pose estimation, and (5) without the posterior update and down-weighting. 

The public evaluation code from \cite{zhou2017unsupervised} was used to obtain our results in Table \ref{table: kitti_odom_results}. The ORB-SLAM results (from \cite{zhou2017unsupervised}) were obtained by running ORB-SLAM \cite{mur2017orb} only on the 5-frame snippets (ORB-SLAM short), and by running ORB-SLAM on the full sequence and chopping the full estimated trajectory into 5-frame snippets (ORB-SLAM full).


\setlength{\belowcaptionskip}{0pt}
\bgroup
\def\arraystretch{1.5} 
\begin{table}
\vspace{6pt}
\caption{Absolute Trajectory Error (ATE RMSE) on the KITTI odometry sequences 09 and 10 computed on 5-frame snippets with the evaluation method in \cite{zhou2017unsupervised} (lower is better). O.M., D.W., and P.U. stand for outlier mask, down-weighting in the pose estimation, and posterior update respectively.}
\centering
	\begin{tabular}{l c c} \hline 
	Method              & Seq. 09                    & Seq. 10           \\ \hline
	ORB-SLAM (full) & 0.014 $\pm$ 0.008 & 0.012 $\pm$ 0.011 \\
	ORB-SLAM (short)            & 0.064 $\pm$ 0.141          & 0.064 $\pm$ 0.130 \\ \hline
	O.M.+P.U.+D.W.  & \textbf{0.060 $\pm$ 0.140} & \textbf{0.035 $\pm$ 0.064} \\ 
	O.M.+D.W.       & 0.063 $\pm$ 0.152          & 0.036 $\pm$ 0.075 \\
	P.U.+D.W.        & 0.077 $\pm$ 0.175          & 0.045 $\pm$ 0.090 \\
	O.M.+P.U.        & 0.075 $\pm$ 0.167          & 0.045 $\pm$ 0.090 \\
	None            & 0.075 $\pm$ 0.168          & 0.044 $\pm$ 0.090 \\ \hline
	\end{tabular}
\label{table: kitti_odom_results}
\end{table}

\setlength{\belowcaptionskip}{-12pt}
\begin{table}
\vspace{6pt}
\caption{Absolute Trajectory Error (ATE RMSE) on four KITTI raw sequences (2011\_09\_26) for the full system and the system without the posterior update and down-weighting.}
\centering
	\begin{tabular}{l c c c c} \hline 
	Method  & 0009 & 0046 & 0059 & 0084           \\ \hline
    Full & \bf{4.09} & \bf{0.59} & \bf{2.78} & \bf{3.35} \\
	None & 4.88 & 0.64 & 2.98 & 3.90 \\ \hline
	\end{tabular}
\label{table: kitti_raw_results}
\vspace{-10pt}
\end{table}


As in Table \ref{table: kitti_odom_results}, the most significant improvement is achieved by using the outlier mask to down-weight the photometric consistency loss (None vs. O.M.+D.W.). This is not surprising since the outlier mask was specifically learned to down-weight pixels to improve pose estimation. The posterior update further brings the error down (O.M.+D.W. vs. O.M.+P.U.+D.W.) through fusing the depth estimates probabilistically and updating the inlier probability for down-weighting. Without the outlier mask prior, the probability model alone (P.U.+D.W.) has limited performance, which demonstrates the importance of imposing such a prior.

Our front-end surpassed the performance of ORB-SLAM (short) in pose estimation. On sequence 09, the three components (O.M., P.U., and D.W.) successfully reduced the error to a level below the ORB-SLAM error. On sequence 10, our front-end is approaching the performance of a full SLAM system which performs joint optimization (bundle adjustment) on multiple keyframes.

We also evaluated the full system compared to the system without the three components on sequences from KITTI raw in Eigen et al.'s test split (excluded during training). $evo$ \cite{grupp2017evo} was used to perform $sim(3)$ alignment with the ground truth trajectory and compute the estimated trajectory RMSE. In Table \ref{table: kitti_raw_results}, the full system outperforms the baseline system, in agreement with the KITTI odometry results (Table \ref{table: kitti_odom_results}).

We provide visualization of the dense point cloud generated by our front-end run on the full sequences 09 and 10 in Fig. \ref{fig: pt_cloud_intro} and Fig. \ref{fig: pt_cloud_results}. In Fig. \ref{fig: pt_cloud_results}, there exists a gap, which should be closed, between the beginning and the end of the trajectory due to the drift accumulated along the trajectory. However, overall the point clouds are visually correct.

\setlength{\belowcaptionskip}{-10pt}
\begin{figure}[t]
\vspace{6pt}
\centering
\begin{center}
    \begin{subfigure}{0.95\linewidth}
	    \centering
		\includegraphics[width=1.0\columnwidth]{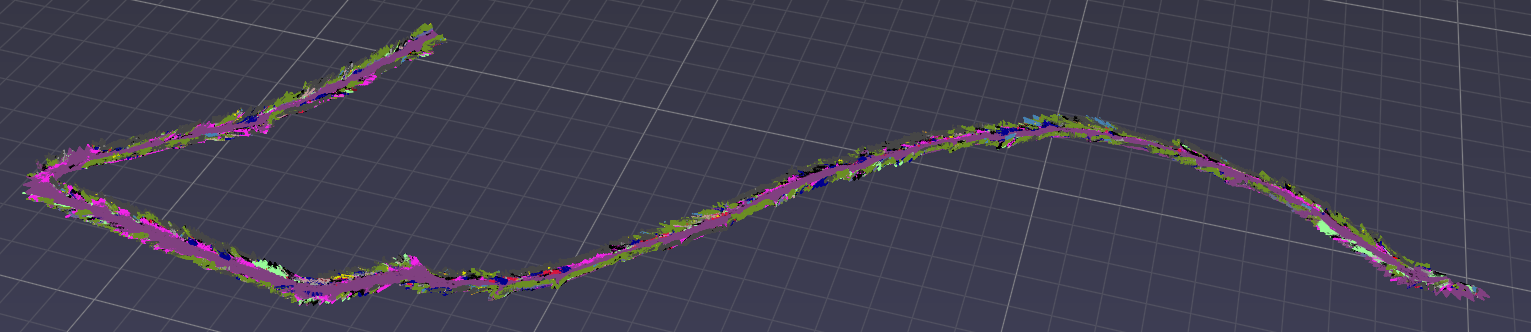}%
	\end{subfigure}%
	\par\smallskip
	\begin{subfigure}{0.95\linewidth}
	    \centering
		\includegraphics[width=1.0\columnwidth]{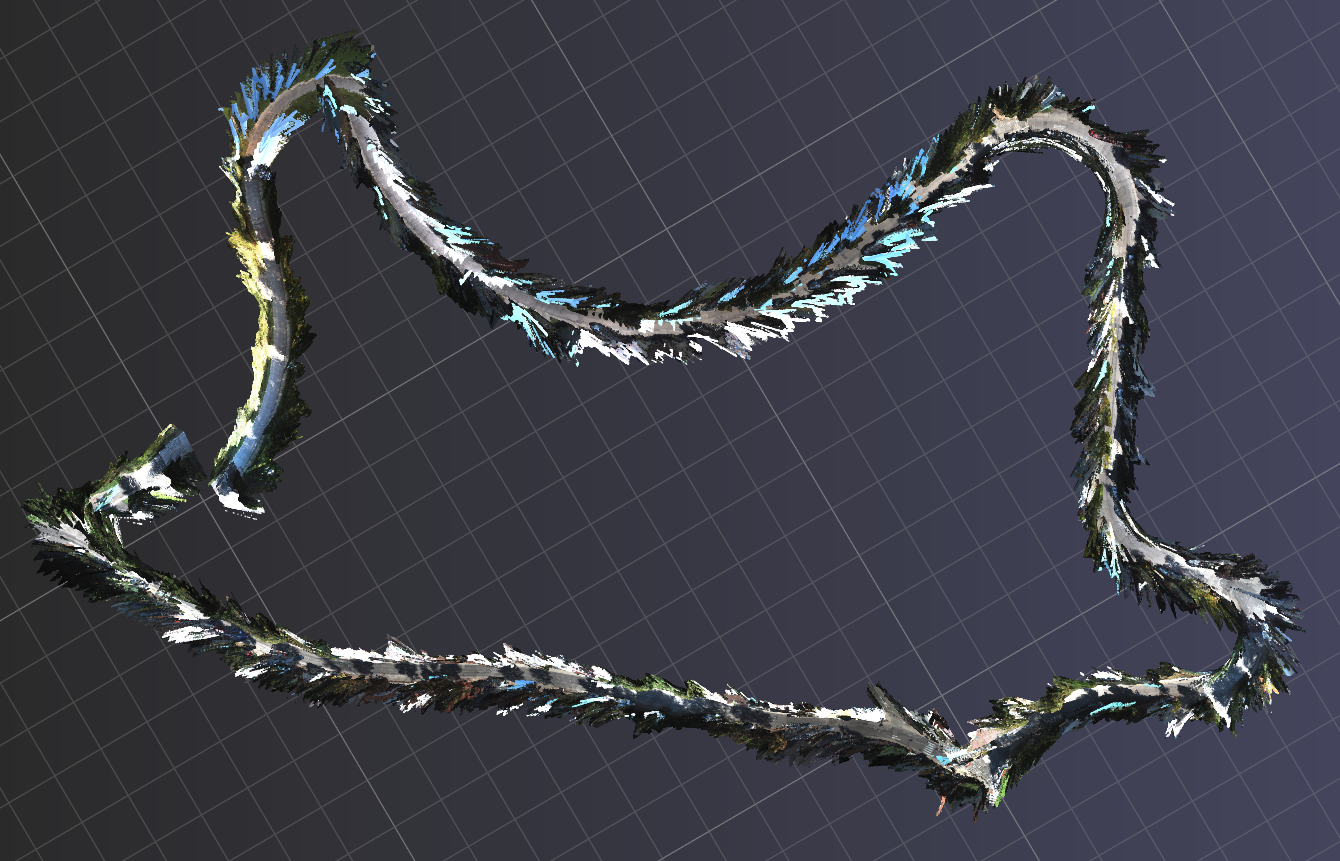}%
	\end{subfigure}%
\end{center}
    \caption{Top row: A semantically labeled dense point cloud of KITTI odometry sequence 10 generated by our front-end. Bottom row: A dense point cloud of sequence 09. (Far points are excluded for cleaner visualization.)}
    \label{fig: pt_cloud_results}
\end{figure}



\subsection{Evaluation on ScanNet}

As in the KITTI case, we followed \cite{zhang2021bootstrapped} to train our depth, ourlier mask, and sementic segmentation predictions on the ScanNet dataset \cite{dai2017scannet}. Four sequences in the validation set (not used during training) were selected for evaluation. We ran the system on 30-frame snippets and computed the mean ATE RMSE for all the estimated poses. Since the snippet is longer, the trajectory evaluation code from \cite{zhou2017unsupervised} was modified to perform trajectory alignment, using the same definition of ATE RMSE as in the TUM evaluation scripts \cite{sturm2012benchmark}, to obtain the results in Table \ref{table: scannet_odom_results}.

\setlength{\belowcaptionskip}{0pt}
\bgroup
\def\arraystretch{1.2} 
\begin{table}
\vspace{6pt}
\caption{Mean Absolute Trajectory Error (ATE RMSE) on 30-frame snippets made from ScanNet \cite{dai2017scannet} sequences. O.M., D.W., and P.U. stand for outlier mask, down-weighting in the pose estimation, and posterior update respectively.}
\centering
	\begin{tabular}{l c c c c c} \hline 
	Method      & 0144\_00         & 0559\_01        & 0565\_00        & 0606\_02        \\ \hline
	Full        & \textbf{0.017}   & \textbf{0.021}  & \textbf{0.018}  &\textbf{0.012}   \\
	O.M.+D.W.   & 0.025            & 0.042           & 0.035           & 0.018           \\
	P.U.+D.W.   & 0.075            & 0.054           & 0.020           & 0.020           \\
	O.M.+P.U.   & 0.081            & 0.024           & 0.019           & 0.015           \\
	None        & 0.018            & 0.027           & 0.021           & 0.026           \\ \hline
	\end{tabular}
\label{table: scannet_odom_results}
\vspace{-10pt}
\end{table}

\bgroup
\def\arraystretch{1.2} 
\begin{table}[t]
\vspace{6pt}
\caption{mIoU of the raw CNN semantic segmentation prediction and the fused semantic labeling through keyframe propagation (\ref{sect: kf_check}).}
\centering
	\begin{tabular}{l c c c c c} \hline 
	              & 0144\_00         & 0559\_01       & 0565\_00        & 0606\_02        \\ \hline
	Raw (\%)      & 16.77            & 11.80           & 15.64           & 16.74           \\
	Fused (\%)    & \textbf{16.82}   & \textbf{11.93}  & \textbf{15.69}  &\textbf{16.80}   \\ \hline
	\end{tabular}
\label{table: scannet_smtic_results}
\end{table}


\setlength{\belowcaptionskip}{-18pt}
\begin{figure}[th]
\centering
	\begin{subfigure}{0.5\linewidth}
	    \centering
		\includegraphics[width=0.88\columnwidth, height=0.88\columnwidth]{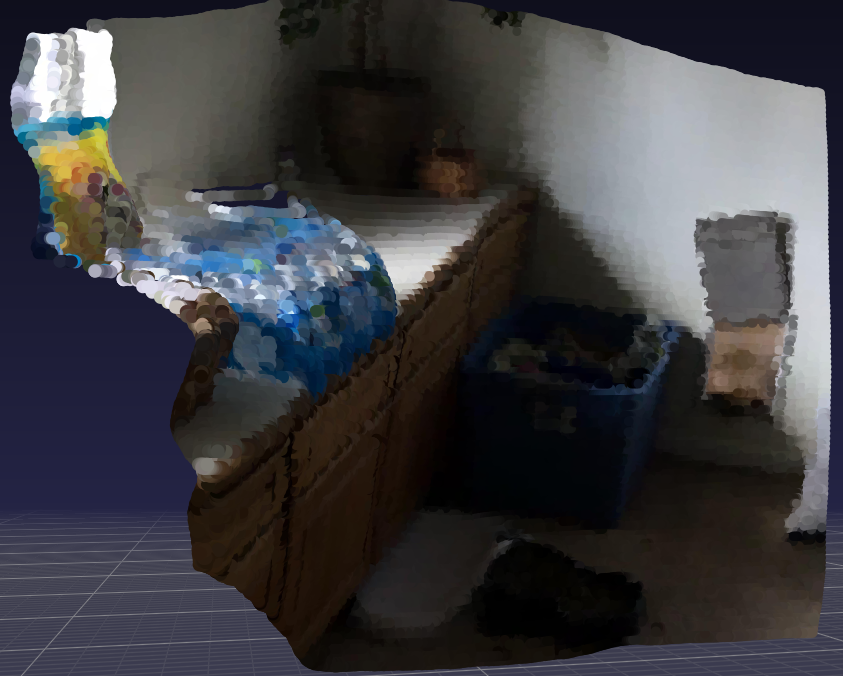}%
	\end{subfigure}%
	\begin{subfigure}{0.5\linewidth}
	    \centering
		\includegraphics[width=0.88\columnwidth, height=0.88\columnwidth]{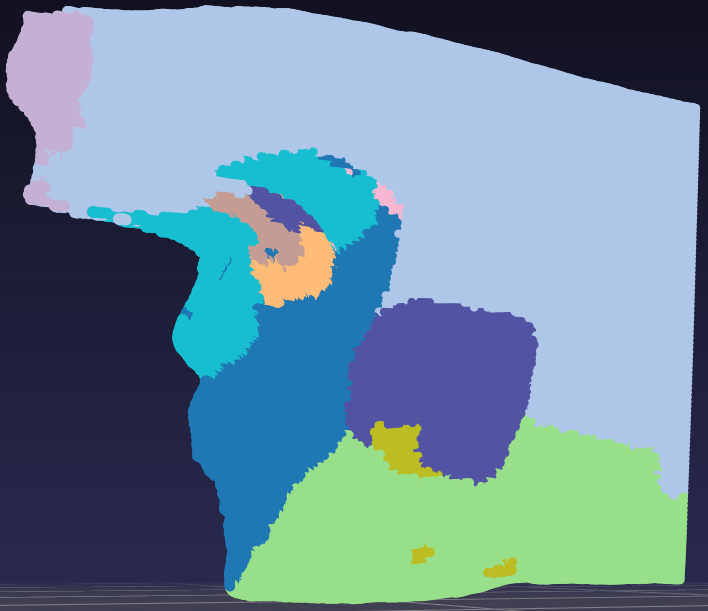}%
	\end{subfigure}%
    \caption{A dense point cloud and its semantically labeled counterpart of a keyframe generated from the ScanNet \cite{dai2017scannet} data by our front-end.}
\label{fig: sn_pt_cloud_results}
\end{figure}

In Table \ref{table: scannet_odom_results}, we observe that the full system consistently performs the best whereas the other down-graded variations have mixed performance. The combination of O.M.+D.W. is less effective than how it was in the KITTI evaluation, possibly because there are no dynamic objects in ScanNet, rendering the outlier mask less useful.  We further look at the fused and raw semantic segmentation images given by the full system in Table \ref{table: scannet_smtic_results}. As shown in the table, our simple keyframe propagation step (\ref{sect: kf_check}) can bring a light improvement to the semantic segmentation. A point cloud generated from a processed keyframe is shown in Fig. \ref{fig: sn_pt_cloud_results}.

\section{Implementation And Performance}

The system parameters were tuned separately for the indoor and outdoor datasets. We found it critical to disable the constant speed motion model (\ref{sect: pose_est}) on ScanNet because it could exaggerate the random and jagged motion of the hand-held camera, occasionally causing the initial point in the pose estimation to leave the converging basin. Other parameters are less critical but adapting them can improve the system efficiency. For example, we can use a shallower image pyramid on ScanNet since the camera motion is small and heavy down-sizing to smooth out the optimization landscape is not needed \cite{engel2013semi}.


All the major computations were implemented with TensorFlow. Inverse depth instead of depth was used in the computations \cite{civera2008inverse}. In our experiments, the system was run on an NVIDIA GTX-1070 GPU at 3 -- 5 Hz depending on the image pyramid and how often the keyframes were inserted. We attribute most of the computation effort to the pose estimation step since the iterative computation of the Hessian in Newton's method is expensive. If we can leverage some better implementation of the direct image alignment method that uses only the Jacobian (e.g. as in Gauss-Newton method), the real-time performance can be greatly improved.

\section{Conclusions}

We have presented a dense CNN-assisted SLAM front-end that can alleviate the outlier effects due to dynamic objects and static occlusions to achieve better pose estimation accuracy. Unlike the traditional approach of using CNN-based object recognition and segmentation to remove dynamic objects and occlusions (e.g. \cite{bescos2018dynaslam}), our method leverages the outlier mask, a by-product from unsupervised learning of depth, which can be more convenient to deploy in a novel environment where only unlabeled data are available. The utility of other outlier-handling techniques used in unsupervised learning of depth (e.g. \cite{godard2019digging}) can be further explored in the context of SLAM in the future.  

\addtolength{\textheight}{-3.5cm}   




\bibliographystyle{IEEEtran}
\bibliography{ICRA2021_zyh}

\end{document}